\documentclass[letterpaper, 10 pt, conference]{ieeeconf}  

\IEEEoverridecommandlockouts                              

\usepackage{times}
\usepackage[bookmarks=true]{hyperref}
\usepackage{graphicx} 
\usepackage{color}
\usepackage{bm}

\usepackage{amsmath, amssymb, amsthm}
\usepackage{mathrsfs}
\usepackage{algorithm}
\usepackage{algorithmicx}
\usepackage{algpseudocode}
\usepackage{gensymb}
\usepackage{cite}
\usepackage{tabu}
\usepackage{multirow}
\usepackage{multicol}
\usepackage{makecell}
\usepackage{booktabs}
\usepackage{xcolor}

\newcommand\Tstrut{\rule{0pt}{2.6ex}}         
\newcommand\Bstrut{\rule[-0.9ex]{0pt}{0pt}}   

\newtheorem*{assumption}{Assumption}
\newtheorem{proposition}{Proposition}

\newcommand{\etal}{\MakeLowercase{\textit{et al.}}}

\title{\LARGE \bf
	Learning Multimodal Confidence for Intention Recognition \\in Human-Robot Interaction
}

\author{Xiyuan Zhao, Huijun Li, Tianyuan Miao, Xianyi Zhu, Zhikai Wei and Aiguo Song 
\thanks{This work was supported in part by the Joint Fund Project 8091B042206 and in part by the National Natural Science Foundation of China under Grant 62173088. \textit{(Corresponding author: Huijun Li.)}}
\thanks{This work involved human subjects or animals in its research. Approval of all ethical and experimental procedures and protocols was granted by the Ethics Committee of Southeast University under Application No. 2020ZDSYLL088P01.}
\thanks{The authors are with the State Key Laboratory of Digital Medical Engineering, Jiangsu Key Laboratory of Remote Measurement and Control, School of Instrument Science and Engineering, Southeast University, Nanjing 210096, China (e-mail: xiyuan.zhao@seu.edu.cn; lihuijun@seu.edu.cn; tymiao@seu.edu.cn; zxianyi@seu.edu.cn; 2302289-33@seu.edu.cn; a.g.song@seu.edu.cn).
}
}

\begin{document}

\maketitle

\begin{abstract}

The rapid development of collaborative robotics has provided a new possibility of helping the elderly who has difficulties in daily life, allowing robots to operate according to specific intentions. However, efficient human-robot cooperation requires natural, accurate and reliable intention recognition in shared environments. The current paramount challenge for this is reducing the uncertainty of multimodal fused intention to be recognized and reasoning adaptively a more reliable result despite current interactive condition. In this work we propose a novel learning-based multimodal fusion framework Batch Multimodal Confidence Learning for Opinion Pool (BMCLOP). 
Our approach combines Bayesian multimodal fusion method and batch confidence learning algorithm to improve accuracy, uncertainty reduction and success rate given the interactive condition.
In particular, the generic and practical multimodal intention recognition framework can be easily extended further. Our desired assistive scenarios consider three modalities gestures, speech and gaze, all of which produce categorical distributions over all the finite intentions.
The proposed method is validated with a six-DoF robot through extensive experiments and exhibits high performance compared to baselines.

\end{abstract}

\smallskip


\section{Introduction}
\label{sec:introduction}

An increased interest and effort in recent years has been paid to human-robot collaboration, most of which is used to support elderly people or assist in space experiments \cite{ajoudani2018progress, xue2020progress}. In such situations, humans are physically incapacitated and in urgent need for robot assistance. Realizing natural and reliable human-robot interaction (HRI) is a crucial part of human-robot collaboration. Numerous modalities were used to realize natural interaction, including gestures, speech, gaze, electromyography (EMG), electroencephalogram (EEG), human skeleton and so on. However, unimodal interaction only conveys limited information, making it difficult to fully express the intention. To guarantee the quality of HRI, it is effective to take advantage of multimodal fusion to attain a better intention. One of the challenges is reducing the uncertainty of fused intentions and reliably reasoning an optimal recognition result given current interactive condition.

From the signal processing point of view, the mainstream approach of multimodal fusion for intention recognition consists of two paradigms \cite{atrey2010multimodal}: feature level and decision level. 
In the feature-level approach, the features extracted from raw data of multiple modalities are first fused and then sent to a common recognition module to perform intention prediction or estimation. 
In this way, the modalities are tightly coupled, which is more suitable for channels that are continuous and synchronized over time, e.g., human skeleton and gestures \cite{roitberg2015multimodal}, hand positions and object features \cite{dutta2019predicting}, skeleton series and environmental features \cite{wei2021vision}. The practical model to fuse feature vectors is neural network, such as Long Short-Term Memory (LSTM) \cite{wei2021vision, doi:10.1177/02783649231210965}, Graph Convolutional Network (GCN) \cite{li2021toward},  
Multimodal Attention Mechanism \cite{magassouba2020multimodal, qian2023gvgnet} etc. Reducing uncertainty and increasing adaptability for such methods is just equivalent to improving the fusion module. In most cases, this requires substantial data to train, which is difficult to acquire in HRI.

In contrast to the previous methods, the decision-level multimodal fusion paradigm is more time-independent, serving to fuse the initial decision results from different modalities. More importantly, unlike the end-to-end fusion models above, the decision-level fusion does not need to be retrained from scratch when modifying modalities. Recent research has also made great progress. Some works \cite{admoni2016predicting, yow2023shared} model the system’s knowledge as a Partially Observable Markov Decision Process (POMDP) where the human intention is the unobservable state. These studies integrate either gaze and joystick using a basic product to attain joint probability distributions or the verbal communication and joystick through question-asking with hindsight optimization. Zhou \etal \cite{zhou2018early} capture the multimodal distributions of EMG, EEG and joint positions to predict turn-taking of robot nurse through Dempster–Shafer theory. Moreover, weighted linear combination of modalities \cite{rodomagoulakis2016multimodal} is also widely used.

\begin{figure*}
	\centering
	\setlength{\abovecaptionskip}{0.1cm}
	\includegraphics[width=0.99\textwidth]{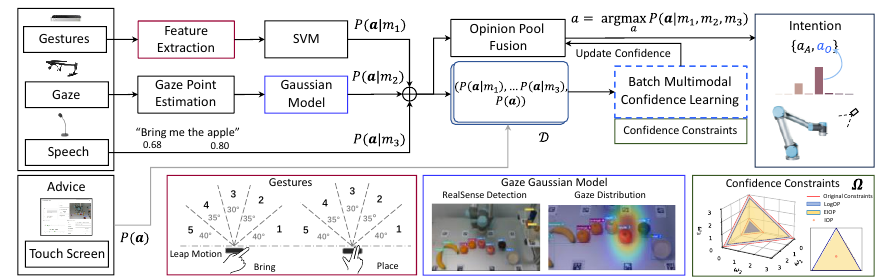}
	\caption{Block diagram of the proposed multimodal fusion framework BMCLOP. When the interactive condition is changed, batch learning with constraints is used to learn the confidence from HRI experiences, in which the deterministic feedback advice as ground truth is given in addition. The object intention is our main focus. 
	$Gestures$: The dashed lines before Leap Motion divide the area into five direction intervals. The bring and place gestures implicitly specify the direction. 
	$Gaze$: We obtain the gaze point in the detected surfaces from Pupil Core and utilize Gaussian distribution to model gaze object distribution.
	$Confidence~Constraints$: We visualize the original and experimental confidence spaces $\bm{\Omega}$ of IOP, LogOP and EIOP respectively.}
	\label{fig:Block_diagram}
\end{figure*}

While several different algorithms in decision level have been proposed, the general consensus to reduce uncertainty is that Bayesian-based method Independent Opinion Pool (IOP) is so far more effective, especially for discrete intention, predicting more accurate and reliable fused results for advantages of reinforcement and mitigation \cite{trick2019multimodal, koert2020multi, trick2022interactive}. The IOP-based fusion method \cite{trick2019multimodal} combines four modalities speech, gestures, gaze directions and scene objects for intention recognition, proving a good experimental result in uncertainty reduction. Other works \cite{koert2020multi}, \cite{trick2022interactive} integrate the IOP fusion for multiple modalities into Interactive Reinforcement Learning by incorporating human advice to accelerate the convergence of reward learning to some extent. However, most methods in decision level implicitly suppose all modalities have the same confidence, i.e., all recognized intentions have the same influence on the fused results. In fact, the modalities have different confidence vectors under different interactive conditions, which is also mentioned in \cite{admoni2016predicting, trick2019multimodal} for further regulation only in discussions.
The sources of uncertainty are not only relevant to the inherent classification algorithm, but also related to modality limitation, external scenarios, task attributes and so on. While IOP-based methods can reinforce the effect of more certain base distributions, it is inherently subject to fixed fusion confidence.
Thus, the development of a framework to quantify modalities' confidence in interactions and guarantee more reliable fusion despite cluttered scenarios is still an open challenge.

To this end, we develop Batch Multimodal Confidence Learning for Opinion Pool (BMCLOP), a learning approach for multimodal fusion, which combines the advantages of Opinion Pool and adaptability to the current interactive condition through constrained optimization. We perform confidence learning when a certain number of interactions are reached. Once the interactive modalities and environment are determined, our approach can ensure the quality of multimodal fusion for intention recognition to be near optimal. It's critical for robot to execute subsequent actions accurately and robustly. Moreover, solving batch learning with constraints is no longer straightforward. Our method is inspired by the work presented by \cite{le2019batch} in reinforcement learning. Note that our work emphasizes multimodal intention recognition instead of prediction. To the best of our knowledge, the study of multimodal confidence learning in HRI is novel.

The contributions of this letter are as follows. With the aim of maximizing the intention recognition performance, we developed a batch learning framework for multimodal fusion based on past experiences in HRI with additionally feedback advice provided by human partner. 

\begin{itemize}
	
	\item We present a fusion framework which learns modalities’ confidence from interactions with only additional human advice. With approximately $32-64$ interactions for batch learning, the approach can learn confidence for adaptive to current interactive condition.
	\item We consider the sources of uncertainty from multimodal human-robot interaction more comprehensively. In addition to uncertainty from base classifiers, we also consider the uncertainty of limited information introduced by the reference of base modalities.
	\item We provide extensive experimental evaluations in robotic kitchen tasks.
	
\end{itemize}

The rest of the letter is as follows: Section \ref{sec:approach} presents the proposed approach. In Section \ref{sec:modalities}, we describe in detail the modalities used to verify the fusion effect. The multimodal fusion framework is evaluated through experiments in desired human-robot interactive scenarios in Section \ref{sec:experiments}. Finally, Section \ref{sec:conclusion} concludes the letter and discusses future research directions.

\section{Proposed Method}
\label{sec:approach}

Our goal is to leverage confidence learned from the past interactions to regulate multimodal fusion process adaptively. Therefore, a learning pipeline for multimodal intention fusion BMCLOP was developed as described by Fig. \ref{fig:Block_diagram}.

Given a natural human-robot interaction task, the uncertainty of intention recognition comprises two key facets. Primarily, the recognition algorithm of each base classifier is inherently influenced by uncertainty. Meanwhile, there is the second source of uncertainty introduced by the limited reference of single modality, which is not considered by previous works \cite{trick2019multimodal, koert2020multi, trick2022interactive}. For instance, speech alone indicating the category of a desired object may be ambiguous when multiple objects of the same category are present. Similarly, a gesture indicating the general direction lacks precision when pointing to a specific object. Combining multiple modalities, the uncertainty of intention can be reduced significantly. 

\subsection{Opinion Pool}

Suppose there are $K$ interactive modalities $m_i~(i = 1, ... , K)$, each of which provides a categorical distribution $P(\bm{a}|m_i)$ for all finite possible intentions $\bm{a}\in\mathcal{A}$. 
For Opinion Pool, we have a paradigm, which combines the multimodal distributions of intentions, given by
\begin{equation}
	P(\bm{a}|m_1,...,m_K) = \alpha(\bm{\omega})\prod_{i=1}^{K} P(\bm{a}|m_i)^{\omega_i} \label{OP}
\end{equation}
where $\alpha(\bm{\omega})$ is the normalization constant. By defining the weight vector $\bm{\omega}$ = $(\omega_1, ..., \omega_K)\in\mathbb{R}^K$ satisfying non-negative, the approach can be used to measure the overall confidence of each modality. The fused distribution using Opinion Pool preserves the multimodal or unimodal nature of the individual base distributions~\cite{genest1986combining}. The most inspiring strength is that not only Opinion Pool fusion is Bayesian, but the influence of each modality on the fused decision can also be adjusted adaptively by changing the confidence vector.

However, due to the multiplicative nature of Opinion Pool, each interactive modality has veto power~\cite{genest1986combining}. Therefore, we enforce the following assumption in case of the above unexpected condition.
\begin{assumption}
	\textup{(Nonzero Probability Property) In this work, all categorical distributions of base modalities are strictly positive each item in the desired intention set $\mathcal{A}$.~\cite{genest1986combining, nedic2017fast}}
\end{assumption}

The commonly used Opinion Pool algorithms contain Independent Opinion Pool (IOP) \cite{trick2019multimodal} and Logarithmic Opinion Pool (LogOP) \cite{genest1986combining, akhondi2014logarithmic, bandyopadhyay2018distributed}. They all have the same fusion structure as \eqref{OP}, but there are different constraints:
\begin{flalign}
	\label{Constraints}
		&\text{IOP:}\qquad\qquad\qquad \omega_1 = ... = \omega_K = 1,& \\
		&\text{LogOP:}\qquad\qquad \sum\nolimits_{i = 1}^{K} \omega_i = 1,~ \bm{\omega} \succeq \bm{0}.&
\end{flalign}
Considering that IOP is fixed, combined with the adaptive advantages of LogOP, we formulate an extended version of IOP (EIOP) as
\begin{equation}
	\sum\nolimits_{i = 1}^{K} \omega_i = K,~ \bm{\omega} \succeq \bm{0}.
\end{equation}
Clearly, IOP is a special case of EIOP where the modalities' weights are equal. Moreover, it's worth mentioning that the revised opinion pool is still Bayesian.

\subsection{Batch Confidence Learning with Constraints}

Under the Assumption above, let $P(\bm{a})$ denotes the deterministic advice for ground truth. The commonly used divergence measure is the Kullback–Leibler (KL) divergence, which measures the degree of similarity between two distributions. The loss function can be defined by
\begin{equation}
	\begin{aligned}
	\mathcal{L}(\bm{\omega})&=\mathbb{E}_{x\sim\mathcal{D}}~D_{KL}(P(\bm{a}) \parallel P(\bm{a}|m_1,...,m_K)) \\
	&=\mathbb{E}_{x\sim\mathcal{D}}~(-H(P(\bm{a})) + H(P(\bm{a}), P(\bm{a}|m_1,...,m_K))) 
	\end{aligned} \notag
\end{equation}
where $x \triangleq (P(\bm{a}|m_1),...,P(\bm{a}|m_K),P(\bm{a}))$ is the recorded human-robot interactive experience sampled from dataset $\mathcal{D}$. We omit $P(\bm{a}|m_1,...,m_K)$’s dependency on weights $\bm{\omega}$ only in notation for the sake of clarity. Here, using the forward KL divergence with respect to $\bm{\omega}$ ensures that the learned model is mode-covering~\cite{bishop2006pattern}. Considering the expression of the KL divergence, we expand the loss function as the sum of an entropy term $-H(P(\bm{a}))$, which does not depend on $\bm{\omega}$, and a cross entropy term $H(P(\bm{a}), P(\bm{a}|m_1,...,m_K))$. 

Thus, we reformulate the learning task as the following optimization problem under constraints
\begin{subequations}
	\begin{align}
		\underset{\bm{\omega}}{\mathrm{min}}~\mathcal{L}(\bm{\omega}) &= \mathbb{E}_{x\sim\mathcal{D}}~ H(P(\bm{a}),P(\bm{a}|m_1,...,m_K)), \\
		\mathrm{s.t.}~\mathcal{G}(\bm{\omega}) &\preceq \bm{0}, \\
		\mathcal{H}(\bm{\omega}) &= \bm{0},
	\end{align}
\label{Constraint}
\end{subequations} 
where $\mathcal{G}(\cdot) = [g_1(\cdot), ..., g_s(\cdot)]^{\mathsf{T}}$ and $s$ is the number of inequality constraints, while $\mathcal{H}(\cdot) = [h_1(\cdot), ..., h_r(\cdot)]^{\mathsf{T}}$ and $r$ is the number of equality constraints. The constraints differ depending on the specific Opinion Pool used above.

\begin{proposition}
	Let $\Omega$ be a convex set of confidence. There is only one fused distribution $P(\bm{a}|m_1,...,m_K)$ to minimize the loss $\mathcal{L}(\bm{\omega})$. \label{proposition1}
\end{proposition}

A brief proof is presented in Appendix. From Proposition \ref{proposition1} and the proof, we know that the loss function is convex. Thus, there exists one or more optimal confidence that performs best in fusing intention recognition.

Generally, we can form an equivalence between the constrained learning and regularized learning through Lagrangian duality, which given by Proposition \ref{proposition2} below.

\begin{proposition}
	Consider the regularized optimization task
	\begin{equation}
		\underset{\bm{\omega}}{\mathrm{min}} ~L(\bm{\omega}, \bm{\lambda}, \bm{\mu}) = \mathcal{L}(\bm{\omega}) + \bm{\lambda}^{\mathsf{T}} \mathcal{G}(\bm{\omega}) + \bm{\mu}^{\mathsf{T}} \mathcal{H}(\bm{\omega})
		\label{Regularization}
	\end{equation}
	Since the strong duality holds in the constrained optimization \eqref{Constraint}, then $\forall \bm{\lambda} \succeq \bm{0}$ such that these two optimization problems share the same optimal solutions.
	\label{proposition2}
\end{proposition}

The Proposition \ref{proposition2} can be derived from Lagrangian duality theory \cite{boyd2004convex}. Obviously, we can learn the current interactive confidence adaptively via solving the optimization problem without constraints. For the reminder of this section, we will describe how BMCLOP performs learning on the limited and small-scale data batch $\mathcal{B}$ in HRI.

\begin{algorithm}
	\caption{BMCLOP.}
	\label{alg1}
	\begin{algorithmic}[1]
		\Require Initial multimodal confidence $\bm{\omega}_{0}$. Batch $\mathcal{B} = \{P(\bm{a}|m_1),...,P(\bm{a}|m_K),P(\bm{a})\}_{i = 1}^{N} \subset \mathcal{D}$. OGD $l_1$ norm bound $B$, learning rate $\eta$. SGD learning rate $\rho$
		\Ensure Learned confidence $\bm{\omega}$
		\State Initialize $\bm{\lambda}_{1} = (\frac{B}{s+1}, ..., \frac{B}{s+1}) \in \mathbb{R}^{s+1}$
		\For{each round $t$}
		\State Learn $\bm{\omega}_t$ $\leftarrow$ SGD$(L(\bm{\omega}, \bm{\lambda}_t), \mathcal{B}, \rho, \bm{\omega}_{0})$
		\State Compute $\hat{\mathcal{L}}(\bm{\omega}_t)$, $\hat{\mathcal{G}}(\bm{\omega}_t)$
		\State $\hat{\bm{\omega}}_t$ $\leftarrow$ $\frac{1}{t} \sum_{t' = 1}^{t} \bm{\omega}_{t'}$
		\State $\hat{\mathcal{L}}(\hat{\bm{\omega}}_t)$ $\leftarrow$ $\frac{1}{t} \sum_{t' = 1}^{t} \hat{\mathcal{L}}(\bm{\omega}_{t'})$, $\hat{\mathcal{G}}(\hat{\bm{\omega}}_t)$ $\leftarrow$ $\frac{1}{t} \sum_{t' = 1}^{t} \hat{\mathcal{G}}(\bm{\omega}_{t'})$
		\State $\hat{\bm{\lambda}}_t$ $\leftarrow$ $\frac{1}{t} \sum_{t' = 1}^{t} \bm{\lambda}_{t'}$
		\State Learn $\tilde{\bm{\omega}}$ $\leftarrow$ SGD$(L(\bm{\omega}, \hat{\bm{\lambda}}_t), \mathcal{B}, \rho, \bm{\omega}_{0})$
		\State Compute $\hat{\mathcal{L}}(\tilde{\bm{\omega}})$, $\hat{\mathcal{G}}(\tilde{\bm{\omega}})$
		\State $\hat{L}_{\max} = \underset{\Vert \bm{\lambda} \Vert_1 = B}{\max}$ $\left( \hat{\mathcal{L}}(\hat{\bm{\omega}}_t) + \bm{\lambda}^{\mathsf{T}} [\hat{\mathcal{G}}(\hat{\bm{\omega}}_t)^{\mathsf{T}}, 0]^{\mathsf{T}} \right)$
		\State $\hat{L}_{\min} = \hat{\mathcal{L}}(\tilde{\bm{\omega}}) + \hat{\bm{\lambda}}_t^{\mathsf{T}} \hat{\mathcal{G}}(\tilde{\bm{\omega}})$
		\If{$\hat{L}_{\max} - \hat{L}_{\min} \leq \varepsilon$}
		\State \Return $\hat{\bm{\omega}}_t$
		\EndIf
		\State $\bm{\lambda}_{t+1}$ $\leftarrow$ EG$(\bm{\lambda}_t, \hat{\mathcal{G}}(\bm{\omega}_t), \eta, B)$
		\EndFor
	\end{algorithmic}
\end{algorithm}

Obviously, the problem \eqref{Regularization} is equivalent to the min-max problem for Lagrangian, which gives by
\begin{equation}
	\underset{\bm{\omega}}{\mathrm{min}} \underset{\bm{\lambda} \succeq \bm{0}, \bm{\mu}}{\mathrm{max}} ~L(\bm{\omega}, \bm{\lambda}, \bm{\mu}).
	\label{Optimization}
\end{equation}
Since the Lagrangian is convex
, strong duality also holds \cite{boyd2004convex}. In practice, equality constraints can be satisfied implicitly by projection in optimization. Thus, we rewrite \eqref{Optimization} as
\begin{equation}
	\underset{\bm{\omega}}{\mathrm{min}}~  \underset{\bm{\lambda} \succeq \bm{0}}{\mathrm{max}} ~L(\bm{\omega}, \bm{\lambda}),
	\label{Optimization_Practice}
\end{equation}
where
\begin{equation}
	L(\bm{\omega}, \bm{\lambda}) = \mathcal{L}(\bm{\omega}) + \bm{\lambda}^{\mathsf{T}} \mathcal{G}(\bm{\omega}).
\end{equation}
To solve this, we analyze the optimization task from the game-theoretic perspective. 
Our task is finding the equilibrium between two sides, one side is $\bm{\omega}-$player, and the other side is $\bm{\lambda}-$player. In every cycle of the game, the $\bm{\omega}-$player minimizes $L(\bm{\omega}, \bm{\lambda})$ given the current $\bm{\lambda}$, the $\bm{\lambda}-$player maximize the Lagrangian given the current $\bm{\omega}$ conversely.

At each iteration, BMCLOP first apply the optimization for $\bm{\omega}$ with regard to $\mathcal{L}(\bm{\omega}) + \bm{\lambda}_{t}^{\mathsf{T}} \mathcal{G}(\bm{\omega})$ given $\bm{\lambda}_{t}$. To achieve a more approximate result to the optimal confidence according to the current batch, a nonlinear minimization procedure should be used to solve the problem, such as Stochastic Gradient Descent (SGD). We start from the given initial value $\bm{\omega}_{t}^{(0)}$ using 
\begin{equation}
	\bm{\omega}_{t}^{(j+1)} = \mathcal{P}_{\Omega} \left( \bm{\omega}_{t}^{(j)} - \rho\nabla L(\bm{\omega}_{t}^{(j)}, \bm{\lambda}_t) \right)
\end{equation}
where $\mathcal{P}_{\Omega}(\cdot)$ is the projection of updated confidence onto the confidence space $\Omega$, $\rho$ is the learning rate of SGD, and stop when either the change $\Vert \bm{\omega}_t^{(j+1)} - \bm{\omega}^{(j)}_t \Vert_{\infty}$ is small enough or a certain number of iterations are exceed.

Next, the $\bm{\lambda}-$player employs online learning approach, which can be any no-regret algorithm that satisfied
\begin{align}
	& \sum_{t} L(\bm{\omega}_t, \bm{\lambda}_t) \geq \underset{\bm{\lambda}}{\mathrm{max}} \sum_{t} L(\bm{\omega}_t, \bm{\lambda}) - o(T).
	\label{RegretCondition}
\end{align}

The most suitable choice is Exponentiated Gradient (EG) algorithm \cite{kivinen1997exponentiated}. Gradient-based methods generally require boundary values to be set. Thus, we introduce hyper-parameters $B$ as the bound of $\bm{\lambda}$ in $l_1$ norm. We also augment $\bm{\lambda}$ into $(m+1)$-dimensional vectors with some abuse of notation. As such, the update step of $\bm{\lambda}$ is given by
\begin{equation}
	\bm{\lambda}_{t+1}[i] = B \frac{\bm{\lambda}_{t}[i] e^{-\eta \bm{z}_t[i]}}{\sum_{j} \bm{\lambda}_{t}[j] e^{-\eta \bm{z}_t[j]}} 
	\label{EG}
\end{equation}
where $\eta$ is the learning rate of EG, $\bm{\lambda}[i]$ is the $i$-th element of the vector $\bm{\lambda}$, and $\bm{z}_t = \left[ \hat{\mathcal{G}}(\bm{\omega}_t)^{\mathsf{T}}, 0 \right]^{\mathsf{T}}$. In the learning process, we optimize the model with two kinds of estimates. The current terms are used to estimate the $\hat{L}_{\max}$ while the expected terms are used to compute $\hat{L}_{\min}$.  After several iterations, the algorithm terminates while the estimated primal-dual gap is less than a given threshold $\varepsilon$. 

Alg. \ref{alg1} presents a pseudo-code of how to learn multimodal confidence from previous interactions. The convergence and close-optimality is guaranteed, which is similar to the detailed analysis in supplementary materials of \cite{le2019batch}.

\subsection{Multimodal Interaction and Learning Procedure}

The multimodal learning process alternates a certain number of interactions with a batch confidence learning procedure, see Fig. \ref{fig:Block_diagram}. The human partner expresses intentions through multiple modalities, which can be recognized by base classifiers. After a data batch is collected, we perform BMCLOP according to Alg. \ref{alg1}. When a certain number of batch learning is executed or the convergence of confidence is achieved, subsequent interactions will not require additional deterministic advice. To make full use of multiple modalities, we consider the lower bound of confidence $\bm{\omega} \succeq 0.1\bm{1}_{K\times1}$ in experiments (see Fig. \ref{fig:Block_diagram}), while learning to strength the effect of better-behaved base distributions.

\section{Considered Modalities}
\label{sec:modalities}

In this study, our primary focus lies on recognizing object intention in complex scenarios, although we also consider two simple actions for experimental completeness. Therefore, we still employ the basic method IOP for action fusion while mainly investigating the effect of our fusion method for object intention. Ambiguity in object reference is infrequent in human-human interactions, largely owing to the innate adaptive multimodal fusion mechanism in humans. 
For intention recognition in human-robot collaboration, the combination of speech and gestures by indicating a general direction is widely used. 
In addition, an implicit gaze is used to track the fixation position for inferring object intention. 
The specific objects are not clear in real-world scenarios, thus we design modalities more generally. The base modalities can be easily replaced by alternative ones. Once this happens, the model confidence should be relearned by Alg. \ref{alg1}.

\subsection{Speech}
For speech recognition, we adopt iFLYTEK, an offline speech recognition wrapped library, which is appropriate for modality reasoning. The user-defined syntax file has to be written to override the recognition range before it can be used.
We define several speech patterns, including two necessary actions (give, place) and five intended object categories (apple, banana, orange, bowl and yogurt). In addition, another words, such as “the” or “me”, which are likely to be used to formulate the speech as a sentence, are also considered. For example, the expression "Give me the apple." is concise but lacks specificity, leading to confusion while there are multiple apples in the scenario. When the speech is recognized, the confidence of each parsed word corresponds to its recognition probability. Assuming uniform probabilities for the remaining words disregarding subtle distinctions, we normalize the output distribution. 
When there are multiple objects of the same category, probabilities are assigned to the same. Subsequently, the distribution is renormalized, and the uncertainty will significantly increase.

\subsection{Gestures}
In our task, we mainly consider two actions, i.e., bring and place. As for the kitchen scene, the gestures “bring” or “place” mean that the human partner asks a robot for help in retrieving the object being pointed at or putting it back. The gestures are illustrated in Fig. \ref{fig:Block_diagram}. To determine the direction of intended object, we also consider directions simultaneously.

We used Leap Motion to recognize gestures and directions. To simplify the task, we divided the area of bring and place gestures into 5 intervals spanning from 0$\degree$ to 180$\degree$, with each representing a general direction. This division was made to mitigate tracking performance issues that arise when the hand is biased towards the edge. It was proved that fingertips angle, fingertips distance and fingertips elevation can effectively extract the hand’s features\cite{marin2014hand}. 
Additionally, the pointing angle (the angle between the projection of direction vector of each finger in the XOZ plane and X-axis) and the component of X-axis for palm coordinate were selected to form a 21-D gesture feature.
Since the Support Vector Machine (SVM) can estimate the distribution without prior, we employed multiclass SVM to provide the base distribution of intention. For training, gestures were recorded from 4 human participants, with each direction being repeated 30 times. To make the model more accurate, C-SVM and RBF kernel were used. The optimal penalty factor and other parameters were determined through traversal to maximize accuracy. 
When multiple objects were positioned in the same direction, equal probabilities were assigned to them before renormalizing distribution of all objects.

\subsection{Gaze}
As an implicit modality, gaze can provide more precise object distribution in most cases. However, due to the lack of action distribution, we only fused the previous two modalities for action recognition. We utilized head-mounted glasses of Pupil Labs to track 2D gaze fixation on the surface specified by markers. After collecting gaze data from Pupil glasses with a rate of 5Hz for 5s, the median and average filtering was adopted to enhance the precision of fixation points. By combining object detection and visual positioning, we transformed the coordinates of fixation and objects into the same coordinate frame. We applied Gaussian distribution $x \sim \mathcal{N}(\mu, \Sigma)$ to model the gaze distribution, where $\mu$ is the filtered fixation point. Since the fixation is more accurate in the horizontal direction, we emphasized the change in the direction. By our experiments, we set the covariance matrix as $\Sigma = \begin{bmatrix} 0.01 & 0 \\ 0 & 0.1 \end{bmatrix}$.

\begin{figure}
	\centering
	\setlength{\abovecaptionskip}{-0.40cm}
	\includegraphics[width=0.99\linewidth]{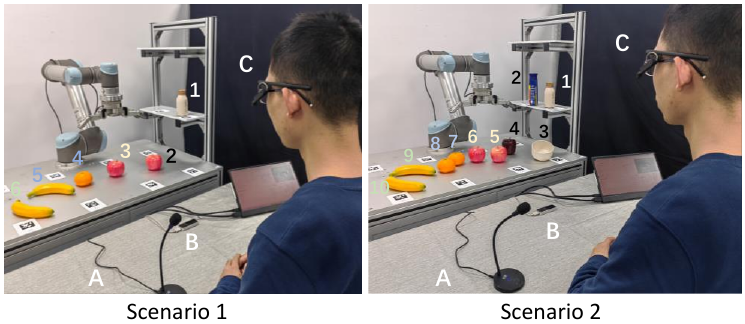}
	\caption{Two robotic kitchen scenarios. We want to prepare some fruit salad, but the objects are all hard to reach. 
	To recognize human's intention, a microphone (A) and Leap Motion (B) are placed on the workspace with head-mounted Pupil glasses (C). Six or ten target objects on the table are object intentions, with each direction colored differently. The touch screen is used for advice input and visual feedback of transparent HRI.}
	\label{fig:Scenario}
\end{figure}

\begin{figure*}
	\centering
	\setlength{\abovecaptionskip}{-0.30cm}
	\includegraphics[width=1.0\textwidth]{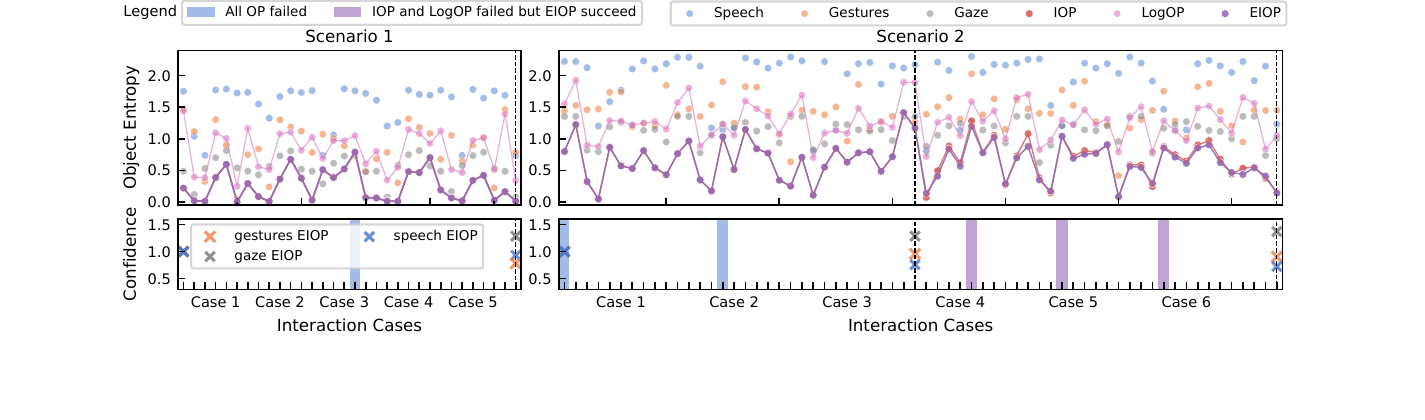}
	\caption{Object entropy and confidence regulation of considered modalities over two kitchen scenarios. Each case involves bringing or replacing objects. The order of interactions is sequential by numbers. After learning from batch, EIOP outperforms all base modalities and the baseline IOP, LogOP.}
	\label{fig:Online_interaction}
\end{figure*}

\begin{figure*}
	\centering
	\setlength{\abovecaptionskip}{0.cm}
	\includegraphics[width=0.98\textwidth]{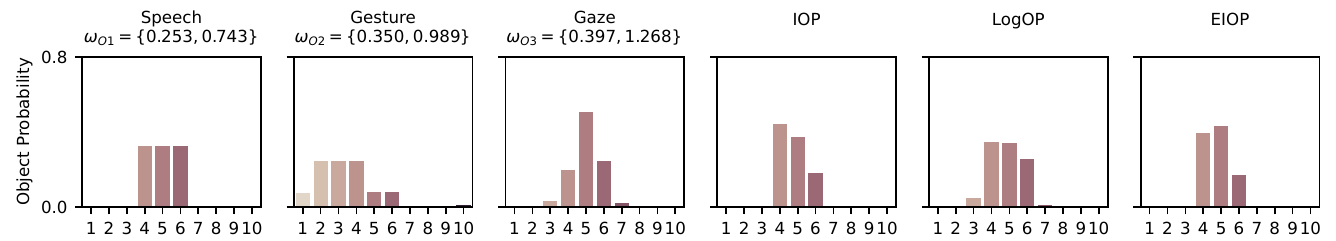}
	\caption{An extreme case of multimodal fusion in the cluttered scenario (Scenario 2). The human intended to take object 5. Due to the high uncertainty from the limited reference of speech and gesture, there are multiple candidates (speech) or large deviation (gesture) with only the correct recognition result of gaze. The fusion of IOP and LogOP failed to strengthen the reliable base distribution, while EIOP achieved the interactive intention satisfactorily. }
	\label{fig:Distribution}
\end{figure*}

\section{Experiments And Evaluations}
\label{sec:experiments}

In this section, the accuracy, uncertainty reduction and regulation to multimodal confidence of our approach are evaluated experimentally in kitchen scenarios. A human partner prepares a fruit salad for dinner 
with the help of a robot manipulator. We need the robot to acquire the human intention through multimodal fusion in order to help bring or place objects, in which the intention is difficult to recognize through base modality. Since the fusion of actions does not involve the second kind of uncertainty we mainly cared about, we only learned and evaluated the multimodal object intentions.

\subsection{Experimental Setup}
The intended objects in experiments include $6$ objects in the simple scenario and $10$ objects in the cluttered scenario. The specific settings are shown in Fig. \ref{fig:Scenario}. The scenarios take full account of two kinds of uncertainty mentioned earlier. The robot deployed in the kitchen scene is a UR5 manipulator equipped with a Barrett Hand as the end-effector.

For all three modalities, the learning process presented in Alg. \ref{alg1} has been implemented in C++ using automatic differentiation library autodiff \cite{autodiff}. The manipulator decomposed collaborative intentions into simple atomic actions to execute, while the robot control and planning were achieved via ROS and MoveIt interface. The classification for gestures was implemented through libSVM \cite{chang2011libsvm}. We used a depth camera Intel Realsense D435i for robotic vision. The object detection algorithm YOLOv7 \cite{wang2023yolov7} was used to detect the scene objects and extract the coordinates of keypoints to grasp. The direction of each object was also determined by visual perception module.
We learned our fusion model using Alg. \ref{alg1} with parameters $\varepsilon = 0.5$, batch size $|\mathcal{B}| = 32$, OGD learning rate $\eta = 0.1$ and bound of lagarange $B = 30$. The initial confidence vectors of LogOP and EIOP were set to $\frac{1}{K}\bm{1}_{K\times1}$ and $\bm{1}_{K\times1}$ respectively. The SGD learning rate was set to $\rho = 0.01$ for LogOP, while it was set to $0.1$ for EIOP.

To evaluate the fusion method, several objective measures are chosen to quantify the accuracy and uncertainty of different distribution, i.e., accuracy, Shannon entropy and score difference \cite{trick2019multimodal}. 
In addition, success rate is also used to assess the overall quality of intention recognition during the past several interactions. It affects the efficiency of human-robot collaboration to a large extent. If the fusion result is always wrong, the efficiency of human-robot collaboration will be greatly reduced.

\begin{table*}[!ht]
	\begin{center}
		\caption{Ablation Study Results of Different Multimodal Object Intention Fusion in Scenario 2. Mean (Standard Deviation) }
		\label{Ablation_table}
		\begin{tabular}{l l c c c c c}
			\hline 
			\multirow{2}{*}{\textbf{Modality}} & \multirow{2}{*}{\textbf{Fusion Method}} &
			\multicolumn{5}{c}{\textbf{Metrics}}\Tstrut\Bstrut \\ &
			& Accuracy $\uparrow$ \Tstrut\Bstrut & Entropy $\downarrow$ & Score Difference $\uparrow$ & Success Rate $(\%)$ $\uparrow$ & Learned Confidence \\
			\hline
			\multicolumn{2}{l}{Speech} \Tstrut\Bstrut & 0.255 (0.107) & 1.856 (0.452) & 0.0286 (0.119) & 46.67 & -- \\
			\multicolumn{2}{l}{Gesture} \Tstrut\Bstrut & 0.310 (0.158) & 1.477 (0.317) & 0.0641 (0.175) & 33.33 & -- \\
			\multicolumn{2}{l}{Gaze} \Tstrut\Bstrut & 0.508 (0.105) & 1.118 (0.207) & 0.195 (0.172) & 88.00 & -- \\
			Speech + Gesture & IOP \Tstrut\Bstrut & 0.466 (0.201) & 1.160 (0.370) & 0.233 (0.266) & 56.00 & -- \\
			& EIOP \Tstrut\Bstrut & 0.465 (0.203) & 1.156 (0.368) & 0.235 (0.268) & 56.00 & 0.969 ~ 1.031 \\
			Speech + Gaze & IOP \Tstrut\Bstrut & 0.606 (0.150) & 0.884 (0.290) & 0.311 (0.256) & \underline{90.67} & -- \\
			& EIOP \Tstrut\Bstrut & 0.618 (0.149) & 0.856 (0.268) & 0.327 (0.261) & \underline{90.67} & 0.902 ~ 1.098 \\
			Gesture + Gaze & IOP \Tstrut\Bstrut & 0.621 (0.186) & 0.808 (0.266) & 0.369 (0.265) & 86.67 & -- \\
			& EIOP \Tstrut\Bstrut & 0.627 (0.183) & 0.799 (0.262) & 0.374 (0.265) & \underline{90.67} & 0.956 ~ 1.044 \\
			Speech + Gesture + Gaze & IOP \Tstrut\Bstrut & \underline{0.687} (0.195) & \underline{0.680} (0.303) & \underline{0.461} (0.303) &\underline{90.67} & -- \\ 
			& EIOP \Tstrut\Bstrut & \textbf{0.715} (0.181) & \textbf{0.630} (0.285) & \textbf{0.499} (0.290) & \textbf{97.33} & 0.725 ~ 0.905 ~ 1.370 \\
			\hline
		\end{tabular}
	\end{center}
\end{table*}

\subsection{Multi-Scenarios Evaluation for Multimodal Fusion}

\begin{figure}
	\centering
	\setlength{\abovecaptionskip}{-0.35cm}
	\includegraphics[width=1.0\linewidth]{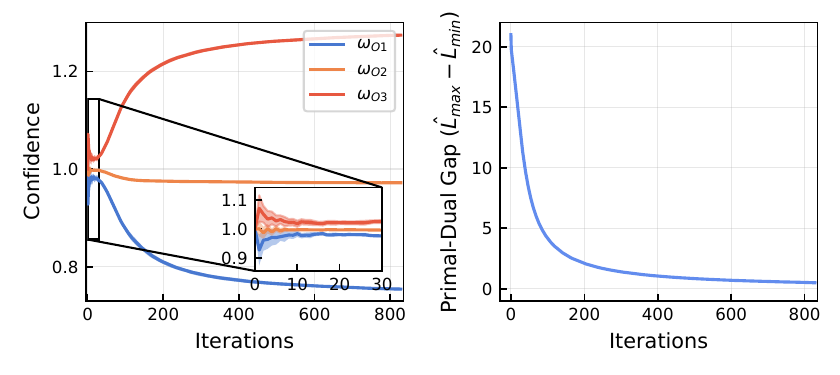}
	\caption{The learning curves of EIOP with the first interactive batch in Scenario 2. Averaged confidence and primal-dual gap across 30 runs are presented over different random seeds. Shadow: 95\% confidence interval. }
	\label{fig:Training}
\end{figure}

We set up two distinct scenarios for comparison: the simple scenario, denoted as Scenario 1, and the cluttered one namely Scenario 2 (Fig. \ref{fig:Scenario}). We used all three modalities to express intentions. After each interaction, multiple modalities as well as advice input were recorded. In both interactions, we performed interactive learning every interactive batch. The primal-dual gap and confidence curves of the initial batch learning for BMCLOP in Scenario 2 are depicted in Fig. \ref{fig:Training}. The main objective is to confirm the adaptive performance of our fusion framework across different scenarios using the same modalities. In contrast, subsequent ablation experiments will allow for the evaluation combining different modalities in the same environment.

See Fig. \ref{fig:Online_interaction} for object entropy and EIOP confidence variation throughout the interaction procedure. Notably, the learned confidence vectors exhibit variations across different scenarios. Considering the differing levels of uncertainty in each scenario, the learning procedure was performed once for Scenario 1 and twice for the cluttered Scenario 2. As learning goes on, the confidence of gaze increases in both scenarios, whereas the speech and gestures exhibit contrasting trends, i.e. $\omega_{O3}>\omega_{O1}>\omega_{O2}$ in Scenario 1, $\omega_{O3}>\omega_{O2}>\omega_{O1}$ in Scenario 2. 
In the cluttered scenario, the uncertainty of reference greatly impacts recognition results of speech and gestures. However, due to the inherent characteristics, the referential uncertainty has the greater effect on speech, while the impact on gesture is relatively small. This explains the relative changes in confidence observed after batch learning in both scenarios. 

For uncertain reduction evaluations, we present the object entropy of considered modalities, as well as three multimodal fusion results during online interaction and learning cases. Each case encompasses six or ten interactions in making salads. Among three modalities, the uncertainty of speech is the highest for multiple objects of the same category. Similarly, when several objects are in the same direction, the uncertainty of gesture also tends to be high. For instance, when pointing to object 4 and 5 in Scenario 1, gestures modality often cause confusion. This is more common in cluttered Scenario 2. 
Throughout the whole interactions, the reduction in uncertainty achieved via IOP and EIOP is remarkable. Although the decision uncertainty of EIOP is marginally lower than that of IOP in most cases, it indicates the effectiveness of our learning framework in ensuring the uncertainty reduction while maintaining adaptability. However, the uncertainty reduction in LogOP is not even as substantial as that achieved in IOP. This suggests that LogOP lacks the ability of reinforcement. From a mathematical perspective, it is evident that the exponent plays a significant role in OP fusion. To this end, the performance of EIOP after batch learning, under equal exponential sum as IOP, has been shown superior uncertainty reduction to varying conditions.

While the enhancement in uncertainty reduction may be somewhat modest compared to IOP, the fusion results of EIOP are more reliable, which guarantees correct recognition results despite extreme conflicting cases, e.g. Fig. \ref{fig:Distribution}. By learning from interactions, our proposed approach could disambiguate among base conflicts and make right decisions (Fig. \ref{fig:Online_interaction}).

\subsection{Confidence Learning Analysis}

For in-depth confidence learning analysis of Alg. \ref{alg1}, we evaluated the learning process in Scenario 2. The algorithm was run for 30 repetitions with a default iterations of 1000 until the gap fell below the threshold $\varepsilon$. We used the first batch of interactive experiences with equal initial confidence.

Results are in Fig. \ref{fig:Training}. During the initial iterations, confidence values oscillate before stabilizing into more gradual changes. The decline in speech confidence $\omega_{O1}$ contrasts with the heightened confidence of gaze $\omega_{O3}$, indicating that the performance of speech diminishes in Scenario 2 due to the uncertainty introduced by multiple candidates. Instead, the referential uncertainty of gaze does not exhibit a sharp increase with the addition of more objects. The gap converges through BMCLOP, highlighting the adaptive nature of EIOP in adjusting interactive conditions based on past interactions.

\subsection{Online Human-Robot Interaction and Ablation Study}

In our online experiments, we performed multimodal intention recognition following several preceding learning iterations in Scenario 2. Notably, no additional advice was required during human-robot interactions. We enlisted three participants to complete the salad-making task 25 times for each person. 
Ablation experiments were conducted to compare fusion performances across different modalities and fusion strategies.

As reported in Table \ref{Ablation_table}, we found that all metrics with IOP and EIOP fusion for three modalities surpassed those from single or dual modality fusion results. Although speech and gestures are less ideal in terms of accuracy and entropy individually, their importance in multimodal fusion stems from the provision of valuable referential information. Following the fusion of speech with gaze or gesture with gaze, the improvements in accuracy, uncertainty reduction and success rate were observed. The sole exception, where a lower success rate was observed, is the combination of gesture and gaze using IOP. The discrepancy is attributed to the high uncertainty of gesture in cluttered scenarios,
where the benefit of information supplement failed to adequately offset the increased uncertainty. 
As can be seen, the performance of BMCLOP achieves a satisfactory fusion behavior in all measures. Compared to baseline IOP, the fusion results demonstrate an enhancement for accuracy of intention recognition while concurrently decreasing the uncertainty of distribution. The success rate of BMCLOP reaches 97.33\%, marking a significant improvement compared with IOP.
The result shows the proposed method exhibits a notable enhancement in success rate for intention fusion, which is particularly important to improve the efficiency of HRI. In simple interactive scenarios, IOP proves to be more convenient and achieves comparable performance to EIOP without requiring additional advice. However, in cluttered scenarios with high uncertainty, IOP or LogOP may fail to recognize the correct intention in certain ambiguous cases. In such instances, EIOP with learned confidence is an indispensable choice.

\section{Conclusion}
\label{sec:conclusion}

In this work, we presented a learning-based fusion approach in the decision level for multimodal intention recognition that works in more complex real-world scenarios with multiple uncertainties. We showed that our method can achieve a better performance in terms of accuracy and uncertainty reduction. To our satisfaction, the success rate is highly increased through batch learning of confidence.

One limitation is that our approach is more suitable for the long-horizon human-robot interactions. Although this work provides an insight for learning multimodal confidence in HRI, 
it still suffers from the matter that can not fine tune online when encountering potential disturbances. In the future, we plan to design a more flexible online learning mechanism. We will also apply our method to multimodal teleoperation to reduce the referential uncertainty in shared manipulations. Moreover, the application of our method to fuse policy and control prior in deploying robotic hybrid control strategy \cite{rana2023bayesian} would be interesting. 

\section*{Appendix}

From \eqref{OP}, the optimization task \eqref{Constraint} is equivalent to:
\begin{equation}
	\underset{\bm{\omega}}{\mathrm{min}}~\mathbb{E}_{x\sim\mathcal{D}}~ \left(-\mathrm{log}\alpha(\bm{\omega}) - \sum_{i = 1}^{K} \omega_{i} \mathbb{E}_{P(\bm{a})} P(\bm{a}|m_i)\right) \label{ProofFunction}
\end{equation}
where 
\begin{equation}
	-\mathrm{log}\alpha(\bm{\omega}) = \mathrm{log}\left( \sum\nolimits_{\bm{a}} \prod_{i = 1}^{K} P(\bm{a}|m_i)^{\omega_i} \right). \notag
\end{equation}

To prove the convexity of the first term, for all $\lambda \in [0, 1]$, we assume that there exist $\bm{\omega}^1$ and $\bm{\omega}^2$ such that
\begin{equation}
	\begin{aligned}
	&-\mathrm{log}\alpha(\lambda\bm{\omega}^1 + (1-\lambda)\bm{\omega}^2) \\
	&= \mathrm{log}\left( \sum\nolimits_{\bm{a}} \left( \prod_{i=1}^{K} P(\bm{a}|m_i)^{\omega_i^1} \right)^{\lambda} \left( \prod_{i=1}^{K} P(\bm{a}|m_i)^{\omega_i^2} \right)^{1-\lambda} \right).
	\end{aligned} \notag
\end{equation}

Considering the Hölder's inequality,
we have
\begin{equation}
	-\mathrm{log}\alpha(\lambda\bm{\omega}^1 + (1-\lambda)\bm{\omega}^2) \leq -(\lambda \mathrm{log}\alpha(\bm{\omega}^1) + (1-\lambda)\alpha(\bm{\omega}^2)). \notag
\end{equation}
Since the second linear term is also convex, the function \eqref{ProofFunction} is convex and has a minimum value.










\bibliographystyle{IEEEtran}
\bibliography{references}




\end{document}